\documentclass[letterpaper, 10 pt, conference]{ieeeconf}  

\IEEEoverridecommandlockouts                              

\title{\LARGE \bf
BayRnTune: Adaptive Bayesian Domain Randomization \\ via Strategic Fine-tuning
}

\author{Tianle Huang$^{1}$ Nitish Sontakke$^{1}$ K. Niranjan Kumar$^{1}$ Irfan Essa$^{1}$\\ Stefanos Nikolaidis$^{2}$ Dennis W. Hong$^{3}$ Sehoon Ha$^{1}$
\thanks{$^{1}$TH, NS, NK, IE, SH are with Georgia Institute of Technology, GA 30332, USA
        {\tt\small \{thuang325, nitishsontakke, niranjankumar, irfan, sehoonha\}@gatech.edu}}%
\thanks{$^{2}$SN is with University of Southern California,  Los Angeles, CA 90007, USA {\tt\small nikolaid@usc.edu}}%
\thanks{$^{3}$ DH is with University of California, Los Angeles, CA 90095, USA {\tt\small dennishong@ucla.edu}}%
}

\usepackage{algorithm}
\usepackage{algpseudocode}
\usepackage{graphicx}

\usepackage{multirow}
\usepackage{amsmath}
\usepackage{amsfonts}
\usepackage{xparse}
\usepackage{xcolor}
\usepackage{flushend}

\NewDocumentCommand{\rsum}{O{} O{} m}{\ensuremath{\sum_{#1}^{#2} \left(#3\right)}}
\NewDocumentCommand{\rsumn}{O{} O{} m}{\ensuremath{\sum_{#1}^{#2} #3}}
\NewDocumentCommand{\rxon}{O{} O{} m}{\ensuremath{\mathop{\mathbb E}_{#1}^{#2} \left[#3\right]}}
\NewDocumentCommand{\rxonn}{O{} O{} m}{\ensuremath{\mathop{\mathbb E}_{#1}^{#2} #3}}
\NewDocumentCommand{\rsup}{O{} O{} m}{\ensuremath{\mathop{\text{sup}}_{#1}^{#2} \left(#3\right)}}
\NewDocumentCommand{\rsupn}{O{} O{} m}{\ensuremath{\mathop{\text{sup}}_{#1}^{#2} #3}}
\NewDocumentCommand{\rargsup}{O{} O{} m}{\ensuremath{\mathop{\text{argsup}}_{#1}^{#2} \left(#3\right)}}
\NewDocumentCommand{\rargsupn}{O{} O{} m}{\ensuremath{\mathop{\text{argsup}}_{#1}^{#2} #3}}
\NewDocumentCommand{\rinf}{O{} O{} m}{\ensuremath{\mathop{\text{inf}}_{#1}^{#2} \left(#3\right)}}
\NewDocumentCommand{\rinfn}{O{} O{} m}{\ensuremath{\mathop{\text{inf}}_{#1}^{#2} #3}}
\NewDocumentCommand{\rarginf}{O{} O{} m}{\ensuremath{\mathop{\text{arginf}}_{#1}^{#2} \left(#3\right)}}
\NewDocumentCommand{\rarginfn}{O{} O{} m}{\ensuremath{\mathop{\text{arginf}}_{#1}^{#2} #3}}
\NewDocumentCommand{\rmax}{O{} O{} m}{\ensuremath{\mathop{\text{max}}_{#1}^{#2} \left(#3\right)}}
\NewDocumentCommand{\rmin}{O{} O{} m}{\ensuremath{\mathop{\text{min}}_{#1}^{#2} \left(#3\right)}}
\NewDocumentCommand{\rmaxn}{O{} O{} m}{\ensuremath{\mathop{\text{max}}_{#1}^{#2} #3}}
\NewDocumentCommand{\rminn}{O{} O{} m}{\ensuremath{\mathop{\text{min}}_{#1}^{#2} #3}}
\NewDocumentCommand{\rargmax}{O{} O{} m}{\ensuremath{\mathop{\text{argmax}}_{#1}^{#2} \left(#3\right)}}
\NewDocumentCommand{\rargmaxn}{O{} O{} m}{\ensuremath{\mathop{\text{argmax}}_{#1}^{#2} #3}}
\NewDocumentCommand{\rargmin}{O{} O{} m}{\ensuremath{\mathop{\text{argmin}}_{#1}^{#2} \left(#3\right)}}
\NewDocumentCommand{\rargminn}{O{} O{} m}{\ensuremath{\mathop{\text{argmin}}_{#1}^{#2} #3}}

\newcommand{\rbr}[1]{\left[#1\right]}
\newcommand{\rcb}[1]{\left\{#1\right\}}

\newcommand{\rnormt}[1]{\left\|#1\right\|_2}

\usepackage{xcolor}
\newcommand{\cmt}[1]{}

\long\def\ignorethis#1{}

\newcommand{\pctab}{\hspace{0.2in}}

\begin{document}

\maketitle
\thispagestyle{empty}
\pagestyle{empty}


\begin{abstract}
Domain randomization (DR), which entails training a policy with randomized dynamics, has proven to be a simple yet effective algorithm for reducing the gap between simulation and the real world. However, DR often requires careful tuning of randomization parameters. Methods like Bayesian Domain Randomization (Bayesian DR) and Active Domain Randomization (Adaptive DR) address this issue by automating parameter range selection using real-world experience. While effective, these algorithms often require long computation time, as a new policy is trained from scratch every iteration. In this work, we propose Adaptive Bayesian Domain Randomization via Strategic Fine-tuning (BayRnTune), which inherits the spirit of BayRn but aims to significantly accelerate the learning processes by fine-tuning from previously learned policy. This idea leads to a critical question: which previous policy should we use as a prior during fine-tuning? We investigated four different fine-tuning strategies and compared them against baseline algorithms in five simulated environments, ranging from simple benchmark tasks to more complex legged robot environments. Our analysis demonstrates that our method yields better rewards in the same amount of timesteps compared to vanilla domain randomization or Bayesian DR.

\end{abstract}

\section{INTRODUCTION}

Deep reinforcement learning (deep RL) has emerged as a promising tool for developing control policies for a wide range of robot control problems, from manipulation to locomotion. However, deep RL trains a policy from a massive amount of simulation experience and transfers it to real robots, often resulting in a serious challenge known as the ``sim-to-real'' gap~\cite{jakobi1995noise}. The difference between simulation and the real world can cause degraded hardware performance or pose risks to the robot and its surroundings. One common approach to reducing the gap is via system identification~\cite{bongard2005nonlinear,gevers2006system}. However, this approach requires prior knowledge from an expert to tune the system parameter spaces.
An alternative technique is domain randomization (DR)~\cite{tobin2017domain}, where an agent is exposed to a diverse set of parameters during training to improve its generalization capabilities. The deep RL community often adopts this technique due to its simplicity and effectiveness over traditional system identification.  

\begin{figure}
    \vspace{-1em}
    \centering
    \includegraphics[width=0.99\columnwidth]{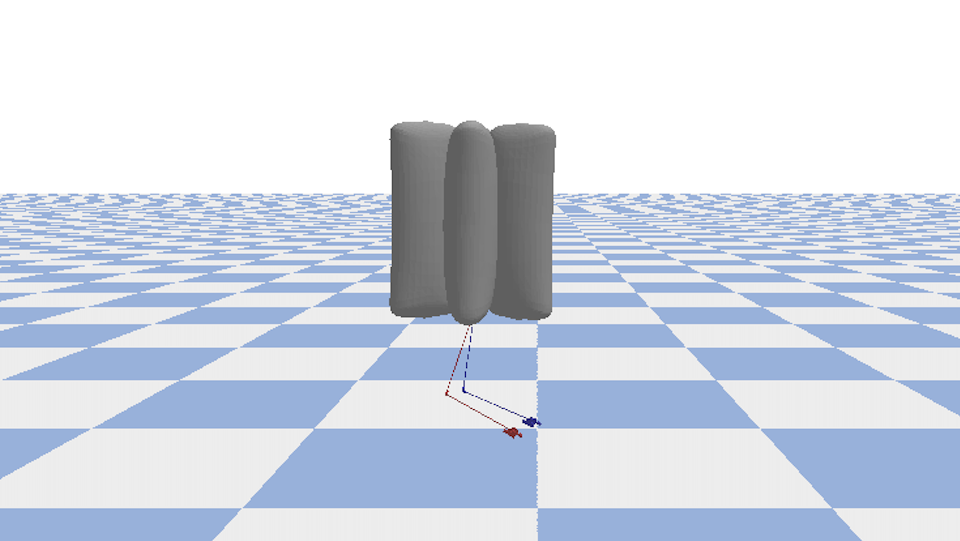}
    \includegraphics[width=0.99\columnwidth]{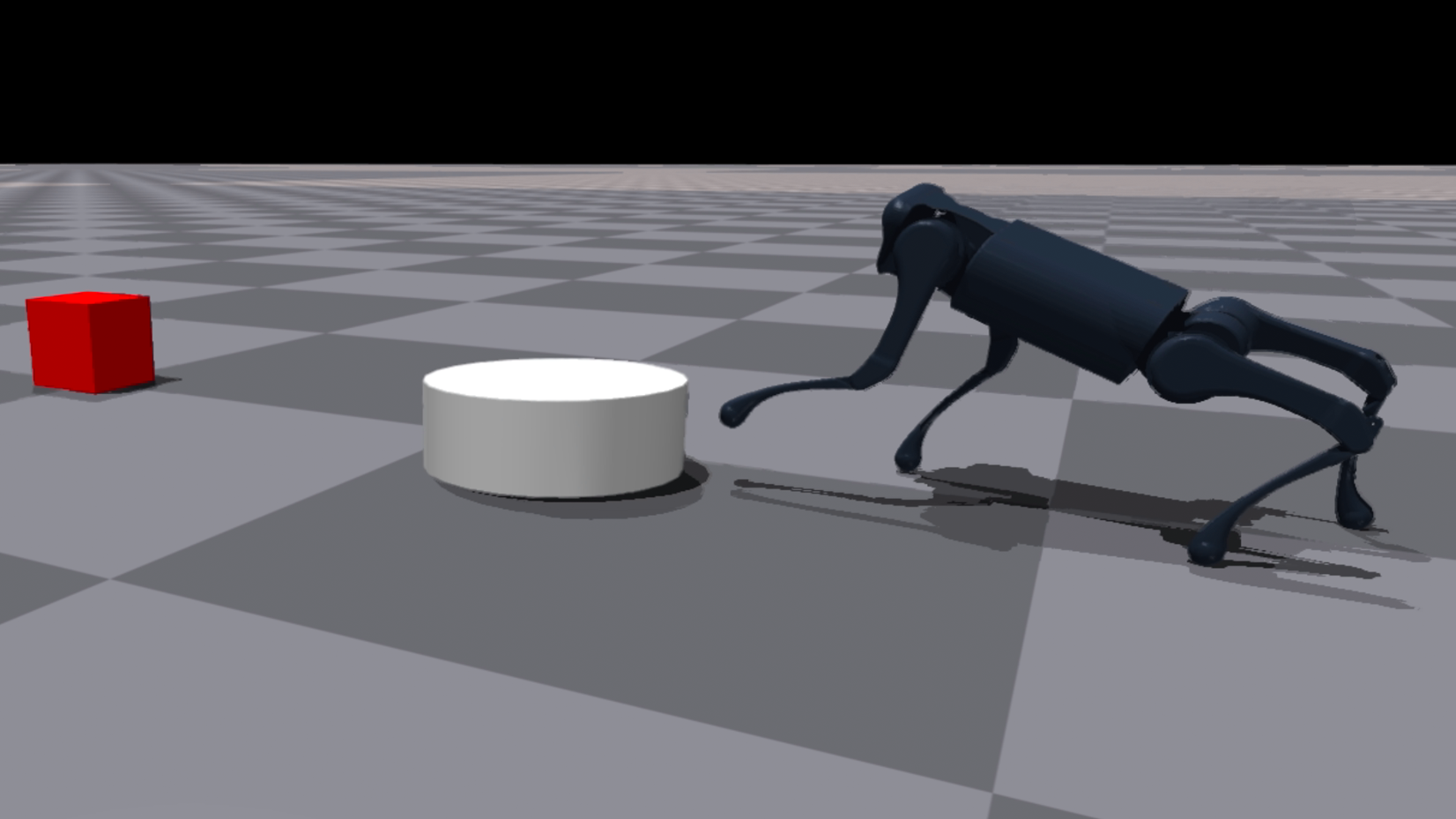}
    \caption{We investigated our novel sim-to-real learning algorithm, BayRnTune, on many robotic tasks, including \emph{BALLU walking} (top) and \emph{Quadruped pushing} (bottom).}
    \label{fig:teaser}
    \vspace{-1em}
\end{figure}


However, determining the appropriate parameters for domain randomization is not straightforward, which has a significant impact on the policy performance. If the range is too narrow, the policy will not be robust to changes in the testing environment. If it is too wide, the learned behavior becomes sub-optimal and conservative~\cite{xie2021dynamics}. To this end, there has been an investigation into automated techniques for adjusting the range of the parameters. Muratore et al.~\cite{muratore2021data} proposed Bayesian DR by combining DR with a gradient-free optimizer, Bayesian Optimization, which adjusts the parameters based on the real-world evaluation. However, this approach requires repetitive policy training with different hyperparameters, which is computationally expensive.

In this work, we propose Adaptive Bayesian Domain Randomization via Strategic Fine-tuning, \emph{BayRnTune}, which is designed to accelerate Bayesian DR by leveraging policy fine-tuning. Our intuition is that the optimal policy for one DR configuration is likely to be close to a previously trained policy and can be reached via fine-tuning, instead of learning from scratch, effectively reducing the training time. However, this strategy raises another follow-up question: What are the best previous policy parameters, or ``checkpoints'', to resume from? We investigate four different tuning strategies, \emph{Normalized Closest Only}, \emph{Infinite Chain}, \emph{Best Only}, and \emph{Best of Last 5}, based on their prioritization of the real-world performance, time-difference, or the distance in the parameter space.

We conducted five simulated sim-to-real experiments on three OpenAI gym benchmark environments (\emph{Hopper}, \emph{HalfCheetah}, and \emph{Ant}) and two robotic control problems (\emph{BALLU walking} and \emph{Quadruped pushing}, Figure~\ref{fig:teaser}). Our results indicate that the proposed BayRnTune performed $25$\% to $400$\% better compared to the baselines of vanilla DR and Bayesian DR, and was able to achieve higher rewards. In addition, we also found that the \emph{Infinite Chain} strategy works the best in most of the problems. We further discuss their convergence, which motivates future research directions.

\section{RELATED WORK}

\label{sec:citations}


\subsection{Sim-to-Real Transfer.}
This paragraph describes related work in sim-to-real transfer with a focus on low-level control. Due to its importance, a variety of techniques have been investigated to mitigate the \emph{sim-to-real} or \emph{reality} gap \cite{jakobi1995noise}. Many prior works focus on improving simulation fidelity by performing system identification~\cite{yu2017preparing, tan2018sim, yu2019sim, hwangbo2019learning, allevato2020tunenet, du2021auto}, establishing the learned models of the environment~\cite{ha2015reducing, yang2020data}, or learning residual dynamics~\cite{golemo2018sim, zeng2020tossingbot, o2022neural, sontakke2023residual}. Alternative approaches involve addressing the differences in data distribution between simulated environments and the real world. This includes domain randomization~\cite{tobin2017domain, peng2018sim, peng2020learning, siekmann2021blind}, adaptation~\cite{peng2020learning, kumar2021rma,chebotar2019closing}, meta-learning~\cite{yu2020learning, o2022neural}, and fine-tuning simulation-trained policies on hardware~\cite{smith2022legged}. There exists another line of research that aims to circumvent the sim-to-real gap by learning policies directly on real robots~\cite{haarnoja2018learning, ha2020learning}. Despite extensive research carried out in this domain, the sim-to-real problem has not been fully solved yet and remains one of the prevalent problems in robotics.


\subsection{Domain Randomization} 
Domain randomization (DR) has emerged as one of the most effective and popular techniques for tackling the sim-to-real problem, particularly in the context of deep RL approaches. The appeal of domain randomization lies in its simplicity and effectiveness in tackling complex phenomena such as mismatch in dynamics, battery discharge, joint slackness, and sensor noise. In addition, DR can also be extended with many techniques mentioned in the previous sections due to its simplicity. These methods have been used to deploy policies successfully on different types of robots, such as manipulators~\cite{tobin2017domain, peng2018sim, akkaya2019solving}, bipedal robots~\cite{siekmann2021blind}, and quadrupedal robots~\cite{peng2020learning}, and many more.

However, one challenge with domain randomization is that it is not straightforward to determine the ranges of the randomization parameters. If it is too narrow, the resulting policy will not be robust. If it is too wide, deep RL will learn a conservative policy that is sub-optimal. Therefore, researchers have proposed systematic approaches to determine the parameters of DR, such as BayesSim~\cite{ramos2019bayessim}, Online BayesSim~\cite{possas2020online}, Automatic DR~\cite{akkaya2019solving}, Bayesian DR~\cite{muratore2021data}, Neural posterior DR \cite{muratore2022neural}, Neural posterior DR~\cite{muratore2022neural} and Active Domain Randomization~\cite{mehta2020active}, which typically leverage real-world evaluation data to calibrate the DR ranges.
While being effective, these techniques can be sample-inefficient because it requires repetitive training in simulation and testing in the real world. On top of this work, we aim to propose a more sample-efficient approach by combining it with fine-tuning. 

\newcommand{\state}{\mathbf{s}}
\newcommand{\sz}{{\state_0}}
\newcommand{\st}{{\state_t}}
\newcommand{\stp}{{\state_{t+1}}}
\newcommand{\action}{\mathbf{a}}
\newcommand{\at}{{\action_t}}
\newcommand{\policy}{\pi}

\section{BACKGROUND}
\noindent \textbf{Deep Reinforcement Learning.}
The task of robot control can be formulated as a Markov Decision Process (MDP). An MDP consists of a state space $\mathcal{S}$, an action space $\mathcal{A}$, a stochastic transition function $p(\stp | \st,\at)$, a reward function $r(\state_t, \action_t)$, and an initial state distribution $p(\state_0)$. By executing a policy $\pi_\theta(\action_t | \st)$, which is parameterized by $\theta$, we can generate a sequence of states and actions known as a trajectory, denoted as $\tau = (\state_0,\action_0, \state_1, \action_1, \ldots)$. The probability of generating a trajectory $\tau$ under the policy $\pi$ is given by $\rho_{\pi_\theta}(\tau)=p(\sz)\prod_t{\pi_\theta}_t(\at|\st)p(\stp|\st,\at)$. Our objective is to find the policy that maximizes the expected sum of rewards over all possible trajectories, as defined by the objective function:

$$
J(\pi_\theta) = \rxon[\tau\sim \rho_{\pi_\theta}]{\sum_{t = 0}^{T} r(\st, \at)}.
$$


\noindent \textbf{Domain Randomization and Bayesian Domain Randomization.}
Domain Randomization (DR) is a widely recognized technique for reducing the sim-to-real gap. Typically, DR ranges are selected by centering them around parameters believed to be close to the actual values of real-world physics. However, two issues arise from this approach. First, users require domain knowledge acquired through calibration or some form of system identification to make an informed selection. Second, even if the correct range is chosen in the simulation, it may not accurately reflect the true dynamics due to underlying modeling assumptions made in the simulator or unmodeled dynamics, such as joint slackness or control delays.

To address the challenge of determining suitable DR parameters, \cite{muratore2021data} propose the Bayesian Domain Randomization (Bayesian DR) algorithm. This algorithm treats the search for DR parameters that maximize the performance in the real world as an optimization problem. Specifically, Bayesian optimization (BO), a gradient-free and sample-efficient algorithm, is employed to identify candidate DR parameters. The real-world performance is obtained by deploying the learned policy and then feeding the results back into the optimizer. By leveraging this feedback mechanism, the algorithm effectively automates the search for favorable DR parameters.



\newcommand{\algname}{BayRnTune\ }

\section{BayRnTune: Bayesian DR with Finetuning}



\subsection{Algorithm}
We introduce the \algname algorithm, which aims to improve the sample efficiency of vanilla Bayesian Domain Randomization via policy fine-tuning. In the Bayesian DR approach, the model is trained from scratch in every optimization step. However, this can be computationally expensive and time-consuming. In contrast, we propose to fine-tune a policy from an existing checkpoint under the assumption that the optimal policy for one DR range is not too far from the optimum for another, and hence, we can improve the sample efficiency drastically.

The \algname Algorithm addresses this inefficiency by preserving all previous checkpoints along with their associated real-world rewards. During each Bayesian optimization step, the algorithm employs a tuning strategy (refer to Section~\ref{tuningStrategies}), which takes the BO-proposed next parameters as inputs and searches the history of all previous optimization steps to determine the appropriate checkpoint to continue the training process from. Once the checkpoint is determined, the learning resumes and continues to fine-tune the given policy.

For each optimization step, we train the model for $T_\text{Tune}$ steps. However, the first optimization step is trained for $T_\text{Bootstrap}$ steps, which is typically much larger than $T_\text{Tune}$ to avoid evaluation of premature policies.
For a detailed description, refer to Algorithm \ref{alg:cap}. 

\begin{algorithm}
\caption{\algname Algorithm}\label{alg:cap}
    \textbf{Input:} 
    Total number of optimization steps $N$; 
    DR parameter range $\boldsymbol\Phi = \rbr{\boldsymbol\phi_{\text{min}}, \boldsymbol\phi_{\text{max}}}$; 
    A parameterized policy $\pi(\cdot)$; 
    Model training algorithm \texttt{PolOpt}$(\theta; \phi; T)$, which trains the model parameterized by $\theta$ in the environment parameterized by $\phi$ for $T$ timestemps; 
    Tuning strategy \texttt {TuneStrat}, see \ref{tuningStrategies} for more details; 
    Oracle for evaluating a model in the real world \texttt{RealEval}. \\
    \textbf{Output:} A learned control policy $\pi(\theta)$.
\begin{algorithmic}[1]
\State Initialize an optimizer \texttt{BayOpt} on space $\boldsymbol\Phi$ 
\State $\theta_0 \gets \texttt{PolOpt}(\theta_\text{rand}; \phi_0; T_\text{Bootstrap})$ \Comment{Initial training}
\State $\mathbf{h} = []$ \Comment{Initialize the history}
\For{$i \gets 1$ to $N$}
    \State $\phi_i \gets$ \texttt{BayOpt.query()}     \Comment{The next DR params}
    \State $\theta^+ \gets$ \texttt{TuneStrat}$(\phi_i; \mathbf{h})$ \Comment{Checkpoint lookup}
    \State $\theta_i \gets$ \texttt{PolOpt}$(\theta^+; \phi_i; T_\text{Tune})$ \Comment{Fine-tuning}
    \State $r_i \gets$ \texttt{RealEval}$(\pi(\theta_i))$ \Comment{Real-world eval}
    \State \texttt{BayOpt.update}$(\phi_i; r_i)$ \Comment{Update BO}

    \State $\mathbf{h}$.append $((\theta_i, \phi_i, r_i))$ 
\EndFor

\State $i^* \gets \rargmaxn[i]{r_i}$ \Comment{Search the best policy}
\State\Return $\pi(\theta_{i^*})$
\end{algorithmic}
\end{algorithm}

\subsection{Tuning strategies} \label{tuningStrategies}


The \algname algorithm is centered around a tuning strategy that plays a crucial role in determining the next steps of optimization. This strategy takes into account several inputs, including the DR parameter for the current optimization step $\phi_i$, the past policies $\theta_1 \dots \theta_{i-1}$, their associated DR parameters $\phi_1, \dots, \phi_{i-1}$, and the real-world rewards $r_1, \dots, r_{i-1}$. Based on these inputs $\mathbf{h}=[(\theta_1, \phi_1, r_1), \dots, (\theta_{i-1}, \phi_{i-1}, r_{i-1})]$, the tuning strategy makes decisions on which policy to train from $\theta^+$. 

In this paper, we have investigated four different tuning strategies: \emph{Normalized Closest Only}, \emph{Infinite Chain}, \emph{Best Only}, and \emph{Best of Last 5}. 




\noindent\textbf{Normalized Closest Only.} The first tuning strategy is \emph{Normalized Closest Only}. It always chooses the policy that is trained with DR parameters that are ``closest" to the current DR parameters. 
To take various dimensions into consideration, we normalize each DR parameter by its standard deviation, and ``closest" is defined to have the smallest 2-norm between the normalized DR parameters. Specifically, we compute the running mean $\boldsymbol\mu$ and standard deviation $\boldsymbol\sigma$ during optimization and get the parameters $\hat\phi = diag(\boldsymbol\sigma)^{-1}(\phi-\boldsymbol\mu)$. 
The checkpoint is then chosen by minimizing the 2-norm of the difference between parameters.

$$
  i^+ = \rargminn[1 \le i' \le i-1]\rnormt{\hat{\phi}_{i, \cdot} - \hat{\phi}_{i', \cdot}},\quad \theta^+ = \theta_{i^+}. \\
$$

\noindent \textbf{Infinite Chain.} In contrast to Normalized Closest Only, the second tuning strategy, \emph{Infinite Chain} simply always chooses to continue to train from the last policy. Intuitively, it is fine tuning one singular model. Specifically,

$$
\begin{aligned}
    \theta^+ &= \theta_{i-1}. \\
\end{aligned}
$$

\noindent\textbf{Best Only.} The third tuning strategy is \emph{Best Only}, which will always continue to train from the policy that produced the highest real world rewards. Specifically,

$$
i^+ = \rargmaxn[1 \le i' \le i - 1]{r_{i'}},\quad \theta^+ = \theta_{i^+}.
$$

\noindent\textbf{Best of Last $M$.} The fourth tuning strategy \emph{Best of Last $M$} chooses the policy that reaches the highest real world rewards from the last $M$ optimization steps. Specifically,

$$
i^+ = \rargmaxn[\rmaxn\rcb{1, i - M} \le i' \le i - 1]{\theta_{i'}}, \quad \theta^+ = \theta_{i^+}. \\
$$

For this study, we specifically choose $M=5$, which yields the \emph{Best of Last 5} strategy.

Note \emph{Best of Last M} can be viewed as a generalization of Infinite Chain and Best Only, as they are special cases where $M = 1$ and $M \rightarrow \infty$, respectively. Choice of $M$ can seen as a balance between exploration vs. exploitation. A larger $M$ encourages taking advantage of previously found good policies, while a smaller $M$ pushes the policy to explore more. 

\section{RESULTS}
\label{sec:result}


In this section, we perform various experiments to analyze the effectiveness of our method. Specifically, we designed our experiments to answer the following three questions.

\begin{enumerate}
    \item Is our \algname more sample-efficient than the baselines, vanilla DR and BayRn?
    \item Which tuning strategy works best for our method?
    \item Do parameters in BayRnTune converge to the ground truth?
\end{enumerate}

\subsection{Task Description}
We evaluated the proposed method in five simulated environments, including three benchmark environments and two robotic tasks.

\noindent \textbf{OpenAI Benchmark.}
Our first three environments are selected from the OpenAI Gym benchmark suite~\cite{brockman2016openai}: Hopper-v3, HalfCheetah-v3, and Ant-v3. All the tasks involve locomotion of the simulated robots either in $2$D (Hopper and HalfCheetah) or $3$D environments (Ant), where the goal is to maximize the distance traveled without falling. For all the experiments, we employ the same BayRnTune hyperparameters, $T_\text{Bootstrap} = 1$M and $T_\text{Tune} = 200$K. Hopper-v3 is trained for $10$M environmental steps, while HalfCheetah-v3 and Ant-v3 are trained for $20$M steps to account for increased complexity. To enable sim-to-real research, we randomize dynamics parameters, such as mass distributions, control frequencies, frictions, and random perturbations as approximation of randomized dynamics in the real world. In the HalfCheetah environment, we introduce the meta parameters of \emph{mass multiplier} and \emph{mass shift} that affect multiple body masses. Particularly, \emph{mass shift} is designed to move COM forward (positive values) or backward (negative values) without changing the robot's total mass. The parameters are summarized in Table~\ref{tab:randomization}.

\begin{table}
\caption{Domain Randomization Parameters}
\vspace{-2em}
\begin{center}
\begin{tabular}{|c|c|c|}
\hline
\textbf{Environment} & \textbf{Parameter} & \textbf{ Range} \\
\hline
Hopper & Per body part mass & $[0, 10]^4$\\
\hline
\multirow{6}{*}{HalfCheetah} & Mass multiplier & $[0.25, 2]$\\
& Mass shift & $[-0.75, 0.75]$\\
& Control delay & $[0, 8]$\\
& Random force change frequency & $[2, 30]$\\
& Random force max magnitude & $[0, 20]$\\
& Friction multiplier & $[0.8, 2]$\\
\hline
\multirow{4}{*}{Ant} & Control delay & $[0, 8]$\\
& Random force change frequency & $[2, 30]$\\
& Random force max magnitude & $[0, 1.25]$\\
& Friction multiplier & $[0.8, 2]$\\
\hline

\multirow{3}{*}{BALLU walking} & Ground friction coefficient & $[0.5, 2.0]$\\
& Drag coefficient & $[0.5, 5.0]$ \\
& External force perturbation & $[-5, 5]$\\
\hline

Quadruped pushing & Puck friction & $[0.5, 1.25]$\\
\hline
\end{tabular}
\label{tab:randomization}
\end{center}
\vspace{-1em}
\end{table}

\noindent \textbf{BALLU Walking.}
Buoyancy Assisted Lightweight Legged Unit (BALLU)~\cite{chae2021ballu2} is a ballon-based legged robot that can walk with its two lightweight legs connected to a base kept afloat using an array of helium-filled balloons. Because of its sensitive dynamics, control of BALLU requires effective sim-to-real transfer. To evaluate our method on this robot, we design a simulated environment using PyBullet~\cite{coumans2016pybullet} similar to the work of Sontakke et al.~\cite{sontakke2023residual}. Here, the task is to walk as quickly as possible while overcoming randomized friction, drag coefficients, and external perturbation forces (Table~\ref{tab:randomization}). We do not randomize the mass and buoyancy parameters because small changes can cause the robot to sink or float away.  We train policies for a total of 32M time steps with $T_\text{Bootstrap} = 6M$ time steps and $T_\text{Tune} = 2M$ time steps. Our reward function consists of two terms that incentivize forward velocity and penalize lateral velocity respectively: $r_t = w_{\textrm{vel}} {v_x}_t - w_{\textrm{penalty}} {v_y}_t$. We set $w_{\textrm{vel}} = 1.0$ and $w_{\textrm{penalty}} = 0.1$ in our experiments.

\noindent \textbf{Quadruped Pushing.}
In this environment, a quadruped robot, Unitree A1, is tasked with moving a cylindrical puck to a given target location. The environment is simulated using NVIDIA IsaacGym~\cite{makoviychuk2021isaac}, a massively parallelizable GPU-based physics simulator. The environment is set up similar to the ``Push object" task in~\cite{kumar2023cascaded}, with a few key differences: the target location is fixed and the surface friction of the puck is randomized to be in the range $[0.5,1.25]$. We choose the true underlying surface friction to be modeled as a truncated normal distribution centered at $0.75$ with a standard deviation of $0.01$. 
We use a reward function and training strategy similar to prior work~\cite{kumar2023cascaded}. 
We define the reward function as a weighted sum of a set of eight terms that control different aspects of the robot motion. We denote the velocity as $\mathbf{v}$, the angular velocity as $\boldsymbol{\omega}$, the joint angles as $\mathbf{q}$, joint velocities as $\dot{\mathbf{q}}$, joint torques as $\boldsymbol{\tau}$, number of robot parts (excluding feet) in contact with the environment as $n\textsubscript{contact}$, action taken at a given step as $a_t$, the position of the target relative to the object being pushed as $\mathbf{x}\textsubscript{t2o}$, and the simulation time-step as $\Delta t$. The reward at time $t$ is defined as the weighted sum of the following quantities:
\begin{enumerate}
    \item Linear velocity tracking:  $\exp(-(\mathbf{v}\textsubscript{target}-\mathbf{v})^2/\sigma_1)$
    \item Angular velocity tracking:  $\exp(-(\omega_{z\textsubscript{target}}-\omega_z)^2/\sigma_2)$
    \item Joint acceleration penalty: $\sum(\frac{\dot{\mathbf{q}}_t-\dot{\mathbf{q}}_{t-1}}{\Delta t})^2$           
    \item Collision penalty: $n\textsubscript{contact}$   
    \item Action change penalty : $\sum(\mathbf{a}_t-\mathbf{a}_{t-1})^2$   
    \item Torque penalty : $\sum \boldsymbol{\tau}^2$
    \item Object-target distance: $\exp(-\left\|\mathbf{x}\textsubscript{t2o} \right\|/\sigma_3 )$
\end{enumerate}
The final reward function is computed by scaling the above terms with weights $[0.01, 0.01,-2.5\times 10^{-7}, -1.0, -0.01, -1.0\times 10^{-4}, 4.0]$ respectively and then adding them together.

We train all our policies as constrained residual perturbations to a pre-learned walking skill. We simulate $4096$ robots in parallel, each initialized with its own puck friction, and train all our policies for a total of $30k$ iterations. The baseline is trained by uniformly sampling the range in Table~\ref{tab:randomization}. The oracle is trained on true parameter distribution directly. Our BO approach models the true parameter as a truncated normal distribution. We train policies for a total of $2.8$B time steps with $T_\text{Bootstrap} = 280$M and $T_\text{Tune} = 280$M.



\subsection{Baseline Algorithms}
\label{subsec:baselines}
We employ several baselines for evaluation.
\begin{itemize}
    \item \textbf{Domain Randomization (Vanilla DR):} This method trains the agent using a large uniform DR range without any adaptation mechanism~\cite{tobin2017domain}.
    \item \textbf{Bayesian DR:} This method adaptively updates the randomization range using Bayesian Optimization, as proposed by Muratore et al.~\cite{muratore2021data}. Each optimization step is trained for $1$M timesteps.
    \item \textbf{Oracle:} For robotic tasks, we also compare our work with an agent trained with the ground-truth dynamics parameters. We intend to use this baseline as an indicator of the upper limits of performance.
\end{itemize}

For the first three OpenAI Gym experiments, we compare our method with all four tuning strategies against the first two baselines, Vanilla DR and Bayesian DR. We use the \emph{Best Only} tuning strategy for the BALLU walking task, whereas for the quadruped pushing task, we selected \emph{Infinite Chain}. We compare these robotic tasks against the Vanilla DR and Oracle baselines.


\subsection{Evaluation}
\begin{figure}
    \centering
    \includegraphics[width=1\columnwidth]{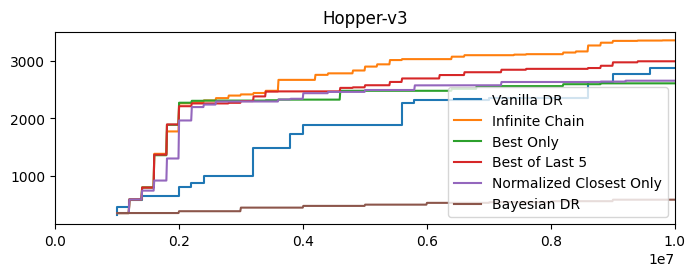}
    \includegraphics[width=1\columnwidth]{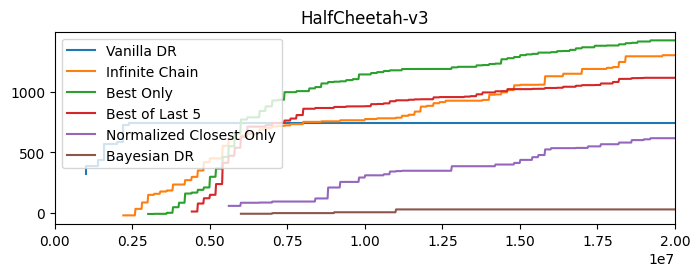}
    \includegraphics[width=1\columnwidth]{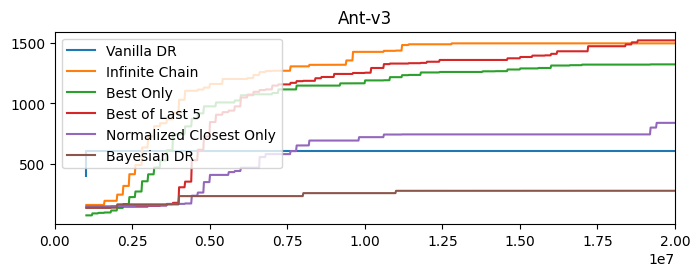}
    \caption{Maximum episodic rewards evaluated in the ground-truth environment. The agents do not have access to the ground-truth parameters. The experiments are repeated with 6 (Ant-v3) or 10 (Hopper-v3 and HalfCheetah-v3) different random seeds and take median values for aggregation. 
    }
    \label{fig:benchmark}
    \vspace{-1em}
\end{figure}

\begin{figure}
    \centering
    \includegraphics[width=1\columnwidth]{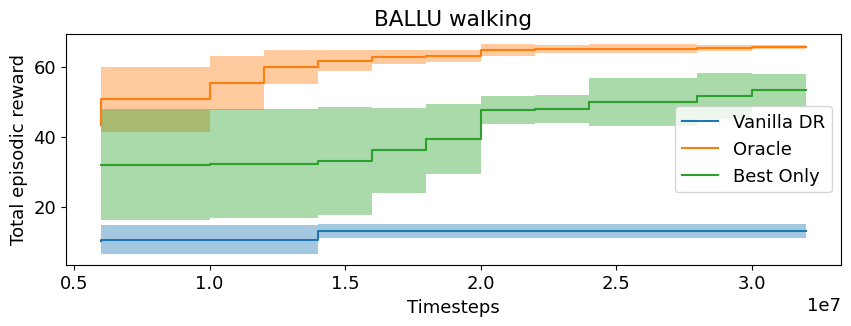}
    \includegraphics[width=1\columnwidth]{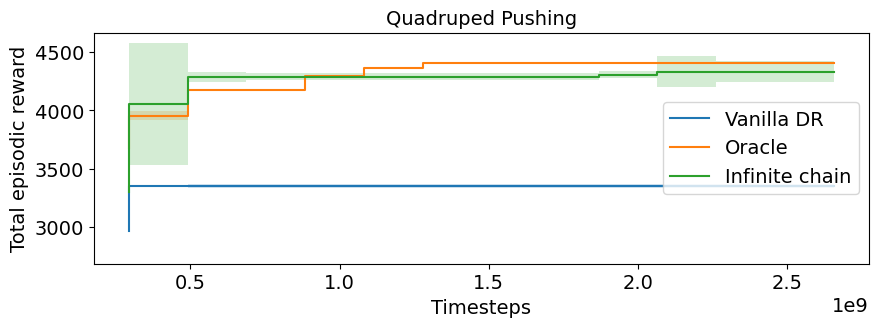}    
    \caption{Robot experiments results. Our BayRnTune (green) outperformed Vanilla DR (blue) and was comparable to Oracle (orange). }
    \label{fig:robot}
\end{figure}

We summarize the performance of agents in the environment with \textbf{ground-truth dynamics} in Figures~\ref{fig:benchmark} and \ref{fig:robot}. Note that the agents do not have access to the ground-truth parameters during training, just like the sim-to-real gap. Instead of standard learning curves, we illustrate the maximum historical episodic reward over time. This aims to consider real-world deployment scenarios where engineers deploy policies regularly and select the best-performing one. Therefore, curves are monotonically increasing over time by definition. We also aggregate multiple experiments by taking their median values in case of benchmark environments to account for outliers and mean for the robotic tasks.

\noindent \textbf{OpenAI Gym Benchmark.}
Overall, three of our strategies (\emph{Infinite Chain}, \emph{Best Only}, and \emph{Best of Last 5}) outperformed both baselines, vanilla DR and Bayesian DR, by significant margins on all three tasks. The \emph{Normalized Closest Only} strategy only worked well in Hopper-v3. Vanilla DR demonstrated reasonable results in all three scenarios, but it often exhibited saturated learning curves due to the lack of adaptation mechanisms. In all experiments, Bayesian DR showed poor performance. However, this is not to say Bayesian DR does not work. The reason it underperforms is that in each optimization step, the policy is only trained for $1$M timesteps and it is not enough to fully saturate the training, which requires about $5$M-$10$M timesteps. On the other hand, all other methods yield the final policy within a total of $10$M or $20$M timesteps.

In our experience, \emph{Infinite Chain} demonstrated the best performance overall: it achieved the highest rewards in Hopper and second-best in the other two environments. In Ant-v3, it was also almost comparable to the top-performing strategy, \emph{Best of Last 5}. The outstanding performance of \emph{Infinite Chain} can be because it trains the same policy for a longer period than the other strategies. In that sense, \emph{Infinite Chain} is somewhat similar to Vanilla DR, except for its adaptation mechanism that greatly boosted its performance. Both \emph{Best Only} and \emph{Best of Last 5} worked reasonably, while the best strategy varied according to the tasks.

Interestingly, \emph{Normalized Closest Only} does not work very well, except for the Hopper environment. This may highlight the multi-modality of control problems: the shorter distance in the randomization parameter space does not necessarily indicate the similarity in the policy parameter space. As a result, we observed that \emph{Normalized Closest Only}  often exhibited premature solutions in many checkpoints.

\noindent \textbf{Robotic Tasks Benchmark.}
\begin{figure}
    \centering
    \includegraphics[width=0.49\columnwidth]{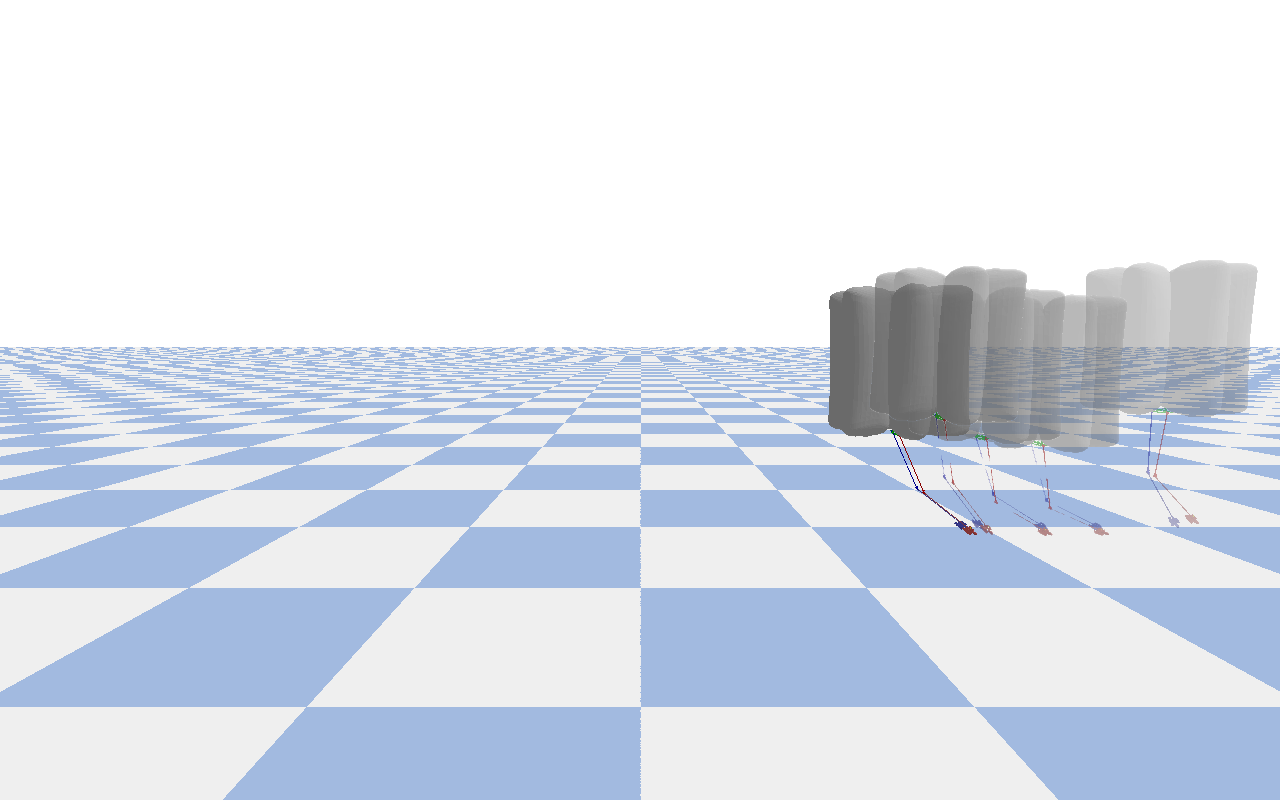}
    \includegraphics[width=0.49\columnwidth]{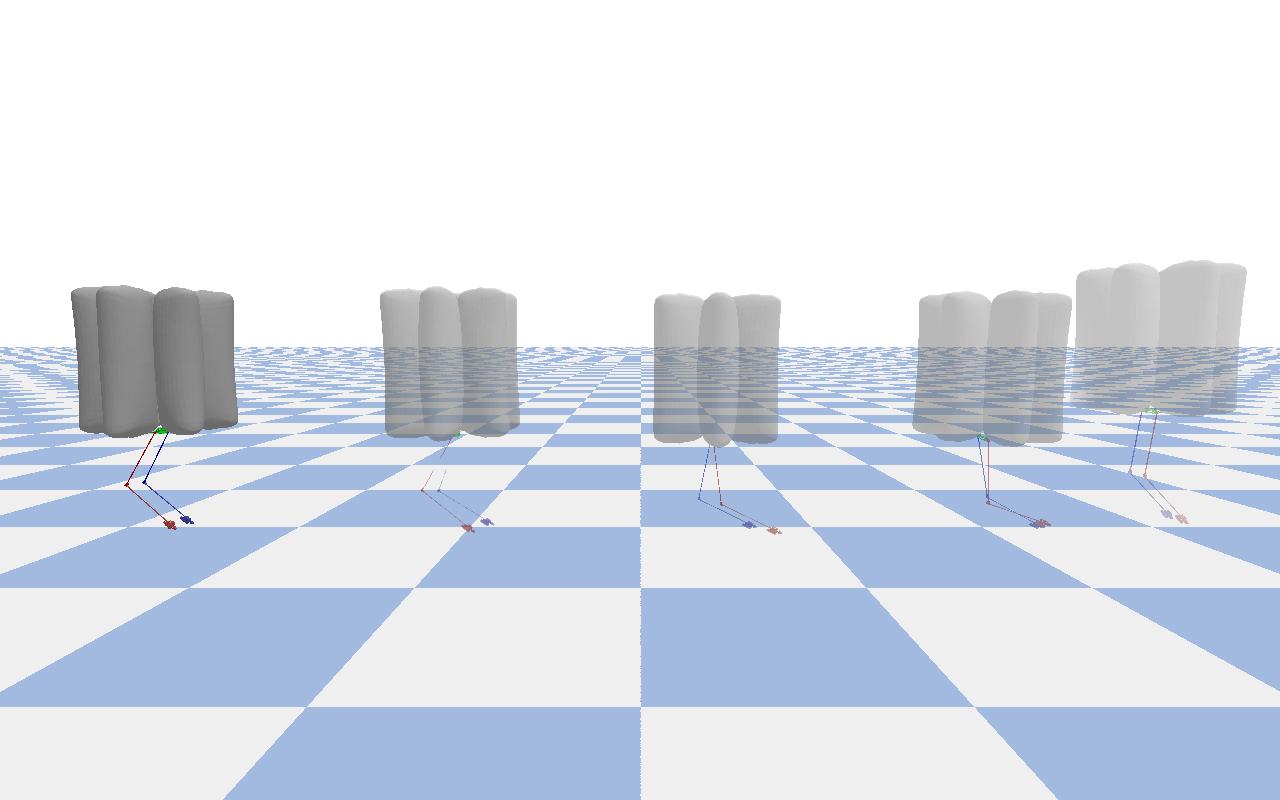}
    \includegraphics[width=0.49\columnwidth]{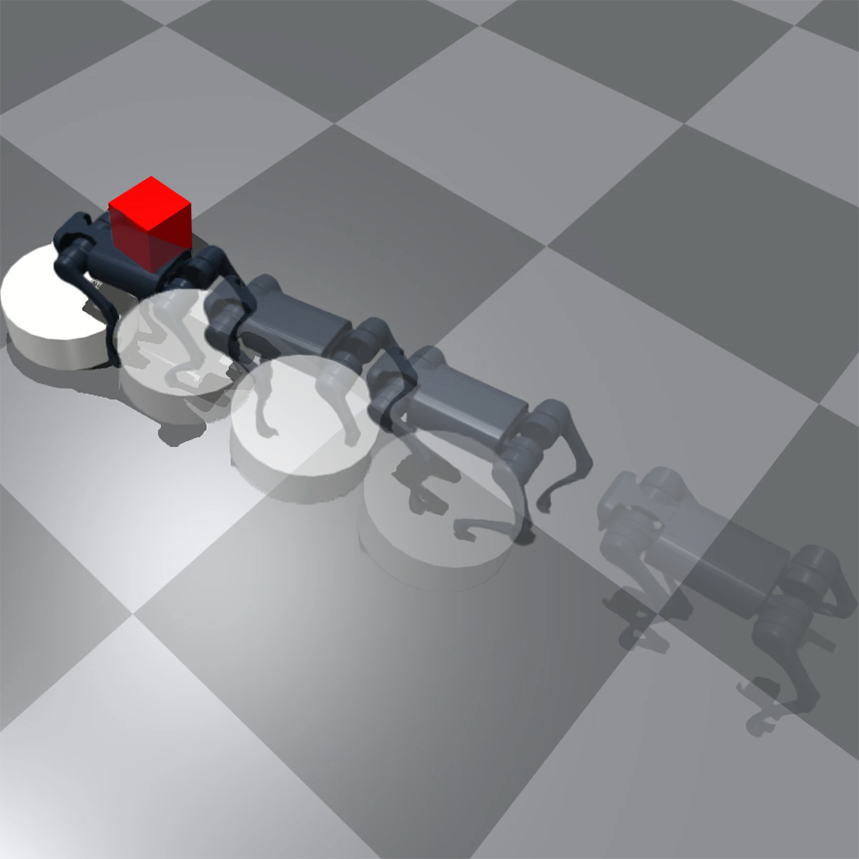}
    \includegraphics[width=0.49\columnwidth]{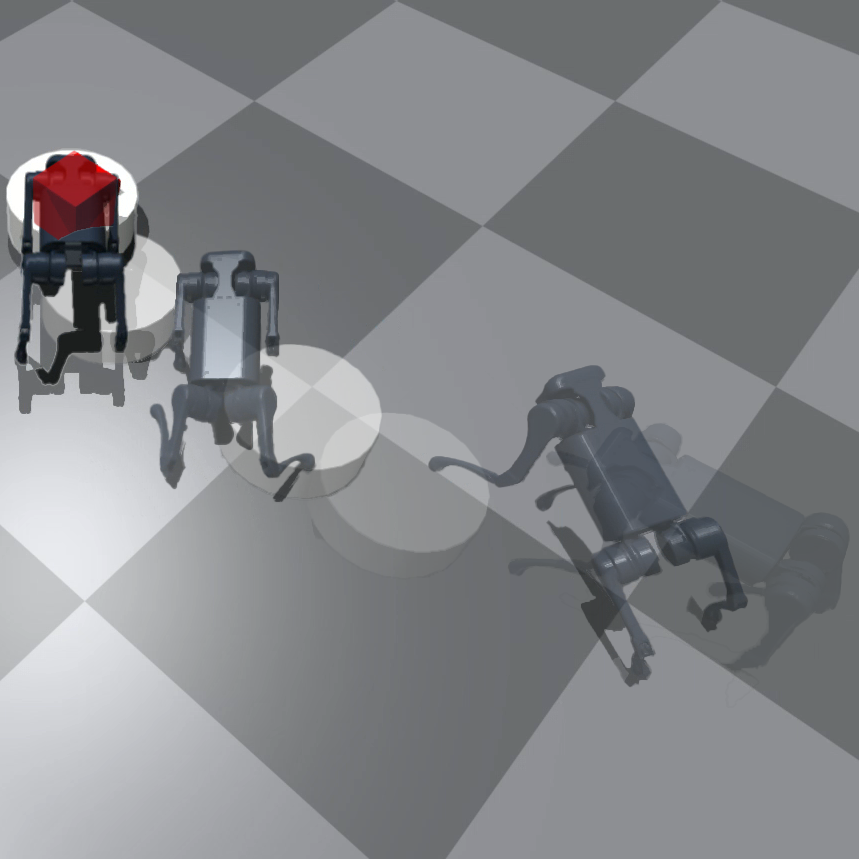}
    \caption{Comparison of Vanilla DR (left) and our BayRnTune (right) on two robotic tasks, BALLU Walking (top) and Quadruped Pushing (bottom).}
    \label{fig:motions}
    \vspace{-1em}
\end{figure}
For the two robotic tasks, we compared our BayRnTune with the \emph{Best Only} and \emph{Infinite Chain} tuning strategies against Vanilla DR and Oracle, as discussed in Section~\ref{subsec:baselines}. For both tasks, our method demonstrated the best performance, comparable to the Oracle trained with ground-truth environment parameters. For the BALLU Walking task, our learned policy achieved a reward of $56.09$ compared to $14.11$ of Vanilla DR, walking $4.19$~m more within $20$ seconds. For the Quadrupedal Pushing task, our agents were able to push the puck $40$~cm closer than the baseline, Vanilla DR, which results in rewards of $4.33k$ and $3.35k$ respectively. The policy trained with our method pushes the puck to the target location almost perfectly, with an average error of $1$~cm. These evaluations indicate that the effectiveness of the proposed BayRnTune is not only limited to simple benchmark environments but also applicable to a wider range of robotic applications. The motions are illustrated in Fig.~\ref{fig:motions} and the supplemental video.

\subsection{Convergence}
\begin{figure}
    \centering
    \includegraphics[width=1\columnwidth]{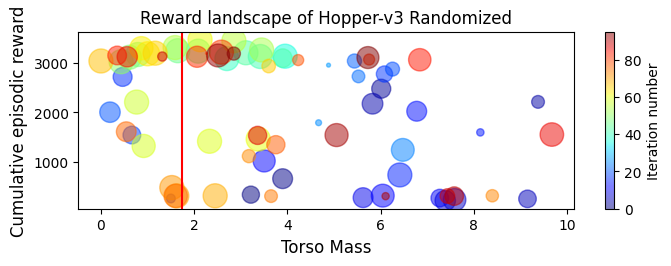}
    \includegraphics[width=\linewidth]{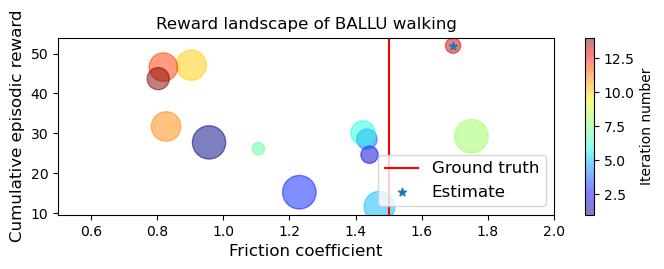}
    \includegraphics[width=1\columnwidth]{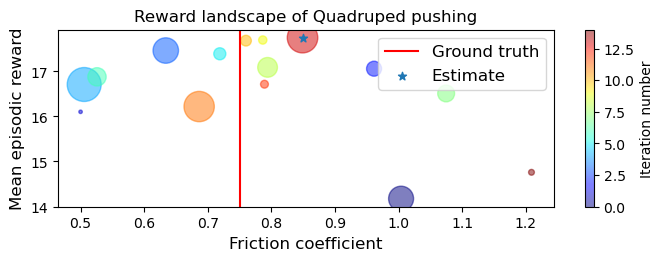}

    \caption{Performance over parameters. While ground-truth parameters lead to the best performance in Hopper (\textbf{first row}) and Quadrupedal pushing (\textbf{third row}) environments, the BO can find another peak at the $1.7$ friction value due to the multi-modality of the problem as in the BALLU walking environment (\textbf{second row}).}
    \label{fig:params}
\end{figure}
Bayesian optimization-based strategies, including the prior work of Bayesian DR and ours, are designed to find the best-performing simulation parameters by maximizing the given black-box objective function. These algorithms are loosely based on the intuition that learning with ground-truth parameters likely results in the maximum episodic rewards. However, it may not be true when the parameters and the performance have non-linear relationships.

Figure~\ref{fig:params} illustrates both cases by plotting the performance over the parameters. The ground-truth parameter is denoted by a vertical line. The sizes of the circles are the associated standard deviations, and the colors represent the iteration numbers. In the Hopper and Quadrupedal pushing environments, we observe clear trends that the deviation from the ground-truth parameters leads to worse performance. For instance, the identified friction of $0.85$ is very similar to the ground-truth friction of $0.75$ in the quadruped pushing environment. On the other hand, in the BALLU walking environment, the BO found another peak performance at the friction of $1.7$ and converged to that parameter. We suspect that its jumping-like policies are easier to learn in high-friction surfaces, which are happen to be robust in a wide range of frictions.

\section{CONCLUSION}
This work presents Adaptive Bayesian Domain Randomization via Strategic Fine-tuning (BayRnTune). Based on Bayesian DR, our algorithm is designed to remove the overhead of repeatedly training policies from scratch and improve sample efficiency by adopting fine-tuning strategies. To determine the starting policy for fine-tuning, we design four strategies, \emph{Normalized Closest Only}, \emph{Infinite Chain}, \emph{Best Only}, and \emph{Best of Last N}, which are based on different philosophies. We evaluated our method on five different tasks, including three OpenAI Gym benchmarks and two robotic tasks on the BALLU robot and the quadrupedal robot. Our BayRnTune significantly outperformed baselines, with the \emph{infinite chain} strategy performing the best on average. We further discussed the convergence of the estimated DR parameters in the analysis.

There exist several interesting future research directions. As discussed in the results section, fine-tuning strategies perform differently in different environments. We believe this is due to the nature of the problem, including the multi-modality of the solution. More rigorous analysis of the effect of reward landscape is another interesting future research direction.To further verify our method’s efficacy, we aim to repeat these experiments on BALLU and A1 hardware in the future.

\section*{Acknowledgement}

This work is supported by the National Science Foundation under Award \#2024768.

\bibliographystyle{ieeetr} 
\bibliography{refs} 

\end{document}